\documentclass[letterpaper, 10 pt, conference]{ieeeconf}  

\IEEEoverridecommandlockouts                              

\usepackage[pdftex]{graphicx}
\usepackage{amssymb}
\usepackage{amsmath}
\usepackage{bm}
\usepackage{cite}
\usepackage{epsfig}
\usepackage{caption}
\usepackage{color}
\usepackage[dvipsnames]{xcolor}
\usepackage{psfrag}
\usepackage{cases}
\usepackage{acronym}
\usepackage{setspace}
\usepackage{times}
\usepackage{latexsym}
\usepackage{dblfloatfix}
\usepackage[ruled,linesnumbered]{algorithm2e}
\usepackage{metalogo}
\usepackage{tabularx}
\usepackage{array}
\usepackage{multirow}
\usepackage{booktabs}
\usepackage{graphicx}
\usepackage{subfigure}
\let\labelindent\relax
\usepackage{enumitem}

\usepackage{hyperref}
 \usepackage{threeparttable}
\hypersetup{
    colorlinks=true,
    linkcolor=blue,
    filecolor=blue,      
    urlcolor=blue,
    citecolor=cyan,
}
\usepackage{geometry}
\title{\LARGE \bf
SSC: Semantic Scan Context for Large-Scale Place Recognition
}

\author{Lin Li$^{1}$, Xin Kong$^{1}$, Xiangrui Zhao$^{1}$, Tianxin Huang$^{1}$ and Yong Liu$^{1,*}$
\thanks{$^{1}$Lin Li, Xin Kong, Xiangrui Zhao, Tianxin Huang and Yong Liu are with the Institute of Cyber-Systems and Control, Zhejiang University, Hangzhou 310027, P. R. China. (*Yong Liu is the corresponding author, email: yongliu@iipc.zju.edu.cn).}
}

\begin{document}

\maketitle
\thispagestyle{empty}
\pagestyle{empty}

\begin{abstract}

Place recognition gives a SLAM system the ability to correct cumulative errors. Unlike images that contain rich texture features, point clouds are almost pure geometric information which makes place recognition based on point clouds challenging. Existing works usually encode low-level features such as coordinate, normal, reflection intensity, etc., as local or global descriptors to represent scenes. Besides, they often ignore the translation between point clouds when matching descriptors. Different from most existing methods, we explore the use of high-level features, namely semantics, to improve the descriptor's representation ability. Also, when matching descriptors, we try to correct the translation between point clouds to improve accuracy. Concretely, we propose a novel global descriptor, Semantic Scan Context, which explores semantic information to represent scenes more effectively. We also present a two-step global semantic ICP to obtain the 3D pose (\(x\), \(y\), \(yaw\)) used to align the point cloud to improve matching performance. Our experiments on the KITTI dataset show that our approach outperforms the state-of-the-art methods with a large margin. Our code is available at: \url{https://github.com/lilin-hitcrt/SSC}.

\end{abstract}

\section{INTRODUCTION}

Simultaneous Localization and Mapping (SLAM) has rapidly developed in recent decades as critical technologies for autonomous vehicles and robots. Place recognition represents the ability of robots to recognize previously visited places, which can build global constraints for the SLAM system to eliminate the odometry's cumulative errors and establish a globally consistent map\cite{4633680}. Place recognition is usually conducted by using images or point clouds. Since point cloud data is rarely affected by environmental factors such as illumination and seasonal changes, LiDAR-based methods have received widespread attention in recent years.

Most existing works on LiDAR-based place recognition are achieved by encoding the point cloud into global or local descriptors and then matching the descriptors. They usually use low-level features such as coordinates\cite{spin,SC,SC1,M2DP,locnet}, normal\cite{ON}, reflection intensity\cite{delight,ISHOT,ISC,ON}, etc. In recent years, with the development of point cloud deep learning, many LiDAR-based object detection\cite{pvrcnn} and semantic segmentation\cite{cylinder3d,spvnas} methods have been proposed, making it possible to obtain semantic information from point clouds. However, there are still only a few LiDAR-based works trying to use semantic information\cite{SGPR,gosmatch,ON}.
  
  \begin{figure}[t]
    \centering
        \includegraphics[width=0.99\columnwidth]{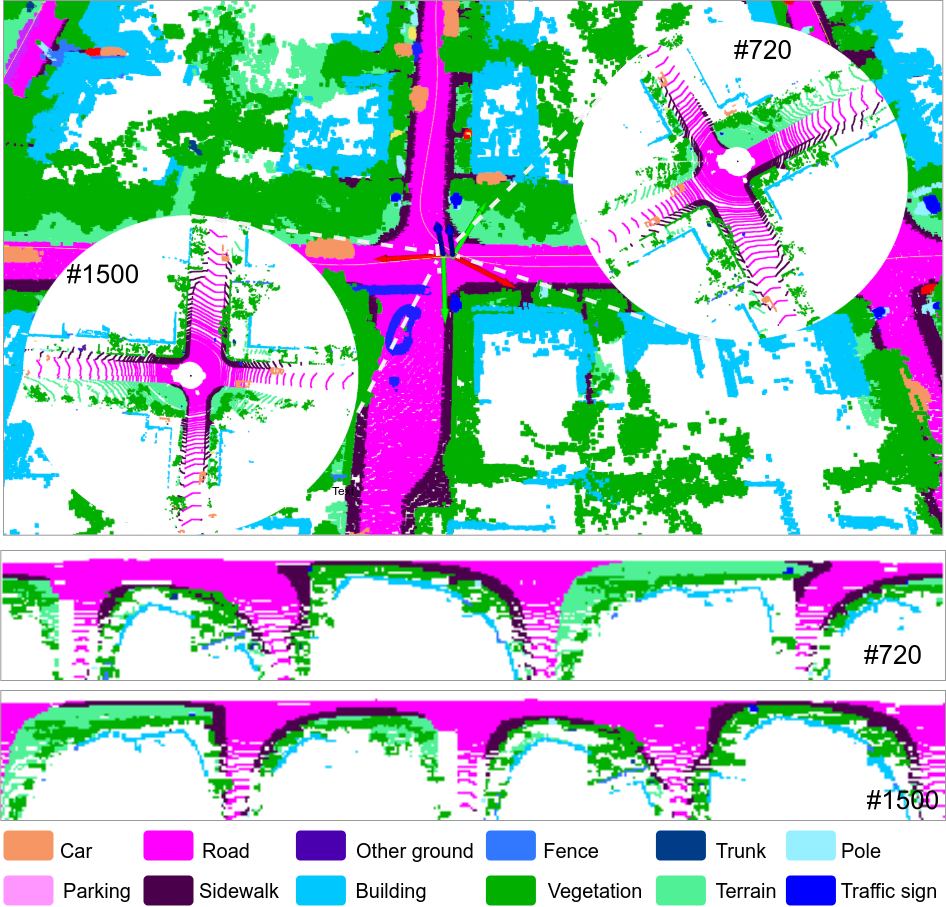}
    \caption{An example of place recognition using semantic scan context. It is a partial map of the KITTI sequence 08, where the frames 720 and 1500 form a reverse loop. The lower part of the figure is the semantic scan context corresponding to the two frames. Since the directions of them are opposite, the descriptors are quite different, while the aligned one shown in Fig.~\ref{pic:pipeline} is easy to distinguish.}
    \label{pic:demo}
 \end{figure}

 \begin{figure*}[t]
    \centering
        \includegraphics[width=2\columnwidth]{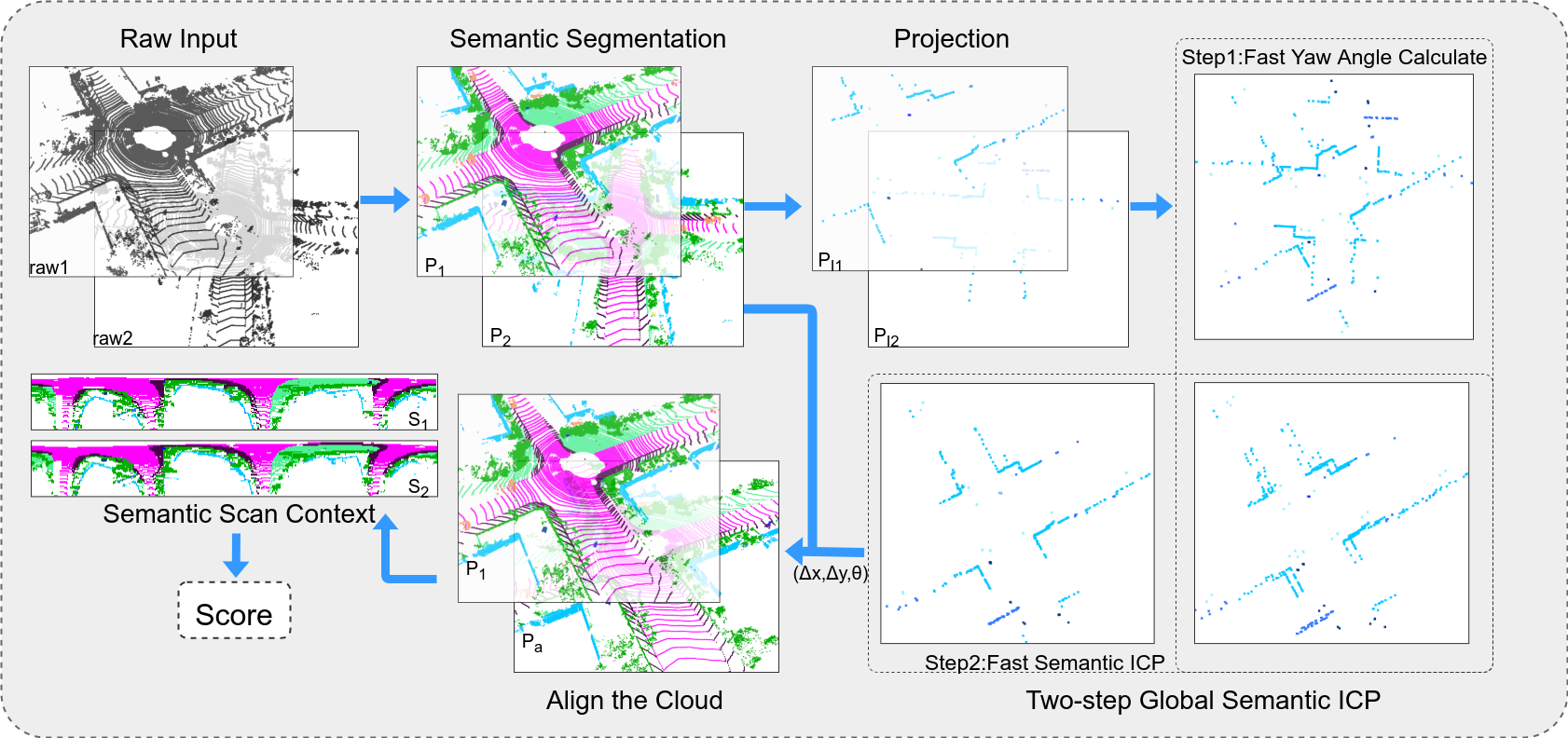}
    \caption{The pipeline of our approach. It mainly consists of two parts: two-step global semantic ICP and Semantic Scan Context. First, we conduct semantic segmentation on the raw point cloud. Then we use semantic information to retain representative objects and project them onto the x-y plane. The two-step global semantic ICP is performed on the projected cloud to get the 3D pose (\(\Delta x,\Delta y,\theta)\). Finally, we use the 3D pose to align the original clouds and generate global descriptors (Semantic Scan Context). The similarity score is obtained by matching SSC.}
    \label{pic:pipeline}
 \end{figure*}

For place recognition, when a robot passes through a place visited before, it does not mean that the two poses are the same. Instead, the robot may walk through the original area from any direction, and there may be a small amount of translation from the original position. Many existing works consider the robot's orientation, namely rotation, and realize the invariance of rotation\cite{SGPR,SC,SC1,ISC}. They may think that the small translation will not strongly impact the recognition result and therefore ignore it. However, we find that simply ignoring the translation for the scan context-based methods will greatly reduce the similarity of the positive samples, making them difficult to identify. 

In this paper, we propose a novel global descriptor named Semantic Scan Context (SSC), which explores semantic information to enhance the expressive power of descriptors. We also propose a two-step global semantic ICP that can produce reliable results regardless of the pose initialization, to obtain the 3D pose $(x,y,yaw)$ of the point cloud. The pose is then used to align the point clouds to reduce the influence of rotation and translation on the similarity of the descriptors. Furthermore, it can also provide good initial values for 6D ICP algorithms to refine the global pose further. Fig.~\ref{pic:demo} is a demonstration of our results. The main contribution is summarized as follows:
\begin{itemize}
    \item We propose a novel global descriptor for LiDAR-based place recognition, which exploits semantic information to encode the 3D scenes effectively.
    \item We propose a two-step global semantic ICP, which doesn't require any initial values, to obtain the 3D pose $(x,y,yaw)$ of the point clouds.
    \item We align point clouds with the obtained 3D poses to eliminate the influence of rotation and translation error on the similarity of the descriptors, which can also further benefit the SLAM system as good initial poses.
    \item Exhaustive experiments on the KITTI odometry dataset show that our approach achieves state-of-the-art performance both in place recognition and pose estimation.
\end{itemize}

\section{RELATED WORK}

According to the features used, we can divide the place recognition methods into three categories: geometry-based, semi-semantic-based, semantic-based.

\textbf{Geometry-based methods}: Spin image~\cite{spin} establishes a local coordinate system for each point, then projects the point into the 2D space and counts the number of points in different areas in the 2D space to form a spin image. ESF~\cite{ESF} proposes a shape descriptor that combines angle, point-distance, and area to boost the recognition rate. M2DP~\cite{M2DP} projects the point cloud into multiple 2D planes and generates a density signature for each plane's points. The left and right singular vectors of those signatures are used as the global descriptors. Scan context~\cite{SC,SC1} converts the point cloud to polar coordinates and then divides it into blocks along the azimuth and radial directions. Lastly, it encodes the z coordinate of the highest point in each block as a 2D global descriptor. LocNet~\cite{locnet} divides a point cloud into rings, generates a distance histogram for each ring, and stitches all histograms to form a global descriptor. Then a siamese network is used to score the similarity between the descriptors. LiDAR Iris~\cite{LI} extracts a binary signature image for each point cloud then uses the Hamming distance of two corresponding binary signature images as the similarity. Seed~\cite{seed} segments the point cloud into different objects and encodes the topological information of the segmented objects into the global descriptor. The above methods have achieved good results by encoding low-level geometric structures into descriptors. It can be expected that integrating more advanced features can further enhance the discriminative power of descriptors.

\textbf{Semi-semantic-based methods}: Some methods use non-geometric information to construct descriptors, such as reflection intensity or learning-based features extracted by neural networks. Such features are related to the object type but do not clearly indicate the semantic category, so we classify these methods as semi-semantic based. ISHOT~\cite{ISHOT} and ISC~\cite{ISC} exploit the intensity information of the point cloud for place recognition. SegMatch~\cite{segmatch} and SegMap~\cite{segmap2} cluster a point cloud into segments. Then they extract features for each segment and use the kNN algorithm to identify corresponds. PointNetVLAD~\cite{PV} combines PointNet~\cite{pointnet} and NetVLAD~\cite{netvlad} to extract global descriptors from the 3D point clouds end-to-end. \(L^3\)-Net~\cite{l3net} selects key points from the given point cloud then uses a PointNet to learn local descriptors for each key point. OREOS~\cite{oreos} projects the 3D point cloud into a 2D range image and proposes a convolutional neural network to extract the global descriptor. DH3D~\cite{dh3d} designs a siamese network to learn 3D local features from the raw 3D point clouds, then use an attention mechanism to aggregate these local features as the global descriptor. LPD-Net~\cite{lpdnet} proposes the adaptive local feature extraction module and the graph-based neighborhood aggregation module to extract local features of the point cloud; then, as the PointNetVLAD, they use the NetVLAD to generate the global descriptor. MinkLoc3D~\cite{minlock3d} uses a sparse voxelized point cloud representation and sparse 3D convolutions to compute a discriminative 3D point cloud descriptor. SeqSphereVLAD~\cite{SeqSphereVLAD} projects the point cloud onto a spherical view, extracts features on it and sequences those features to form a descriptor. SpoxelNet~\cite{spoxenet} voxelized the point cloud in spherical coordinates and defines the occupancy of each voxel in ternary values. Then they use a neural network to extract the global descriptor. The above methods combine more advanced features with geometric features. However, most of them use neural networks to extract abstract features, which are more complicated and not well interpretable.

\textbf{Semantic-based methods}: SGPR~\cite{SGPR} represents the scene as a semantic graph then uses a graph similarity network to score the similarity of the graphs. GOSMatch~\cite{gosmatch} proposes a new global descriptor that is generated from the spatial relationship between semantics. It also proposes a coarse-to-fine strategy to efficiently search loop closures and gives an accurate 6-DOF initial pose estimation. The two methods represent the scene as a graph and abstract the object as a node in the graph, which will cause the loss of features such as the size of each object. OverlapNet~\cite{ON} designed a deep neural network that uses different types of information, such as intensity, normal, and semantics generated from LiDAR scans, to provide overlap and relative yaw angle estimates between paired 3D scans. However, it is too slow in preprocessing due to the need to calculating the normal and inferring the complex network backbone. To use the semantic information more effectively, we propose our Semantic Scan Context approach.

\section{METHODOLOGY}

In this section, we present our semantic scan context approach. Different from other scan context-based methods that use incomplete semantic information and ignore small translations between point clouds, we explore to exploit full semantic information and emphasize that the small translation between point cloud pairs has a significant influence on the accuracy of recognition.

As shown in Fig.~\ref{pic:pipeline}, our method consists of two main parts: two-step global semantic ICP and Semantic Scan Context. The two-step global semantic ICP is divided into Fast Yaw Angle Calculate and Fast Semantic ICP. First, we define a point cloud frame as \( P=\{ p_1,p_2,\cdots,p_n\}\), with each point \( p_i=[x_i,y_i,z_i,\eta_i]\), \(\eta_i\) represent the semantic label of \(p_i\). Given a pair of point clouds \( (P_1, P_2)\), we first use our Fast Yaw Angle Calculate method to get the relative yaw angle \( \theta \) between them. Then we use the Fast Semantic ICP to calculate their relative translation \( (\Delta x, \Delta y)\) in the x-y plane. Through the above two steps, we get the relative poses \( (\Delta x, \Delta y, \theta)\) of the two frames of point clouds in 3D pose space. In order to eliminate the influence of rotation (e.g., reverse loop closures) and small translation on recognition, we use the obtained relative pose to align point cloud \( P_2\). We mark the aligned point cloud as \(P_{a}\). Finally, we use our global descriptor -- the Semantic Scan Context to describe \( (P_1, P_{a})\) as \( (S_1, S_2)\). The similarity score is obtained by comparing \( S_1\) and \(S_2\).

\begin{figure}[t]
    \centering
        \subfigure[Yaw Aligned]{
        \includegraphics[width=0.46\columnwidth]{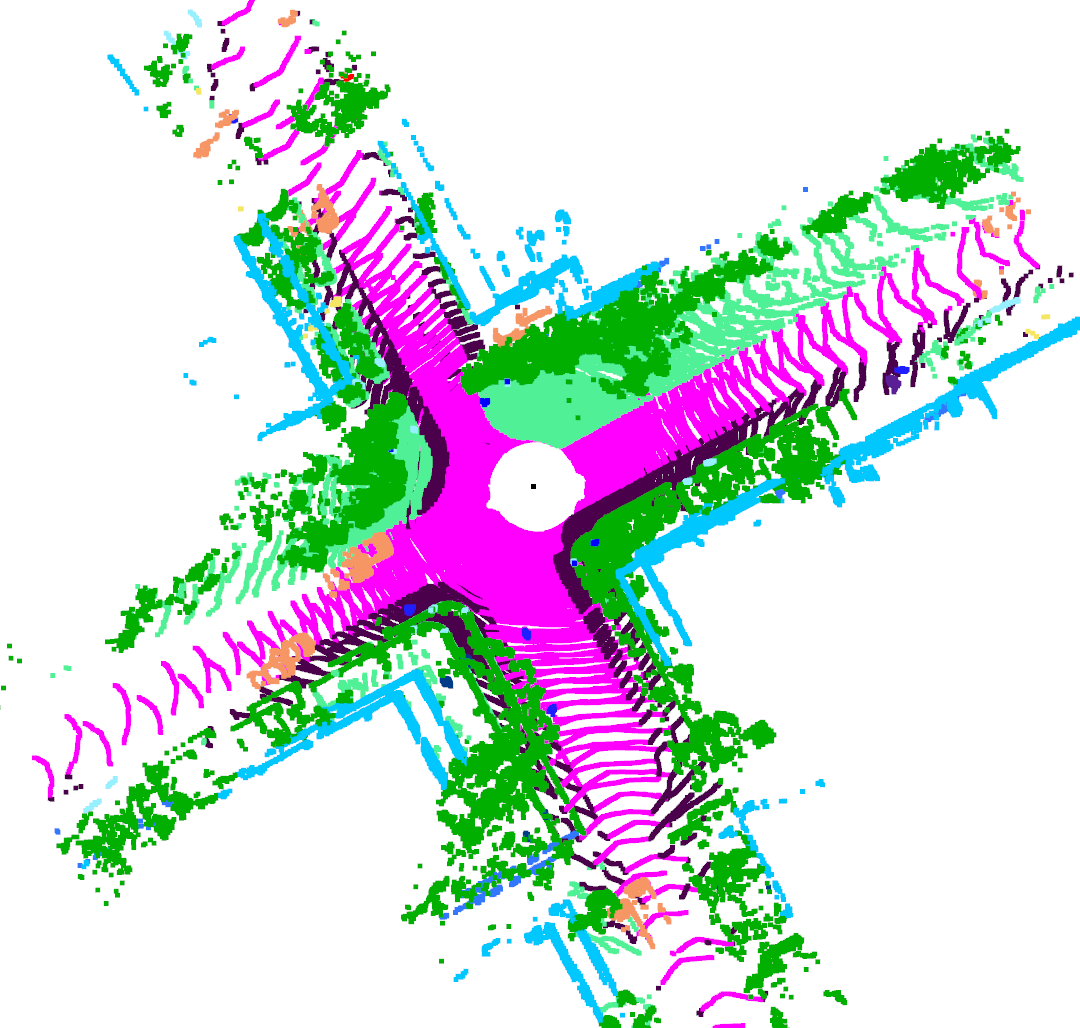}
        }
        \subfigure[Translation Aligned]{
        \includegraphics[width=0.46\columnwidth]{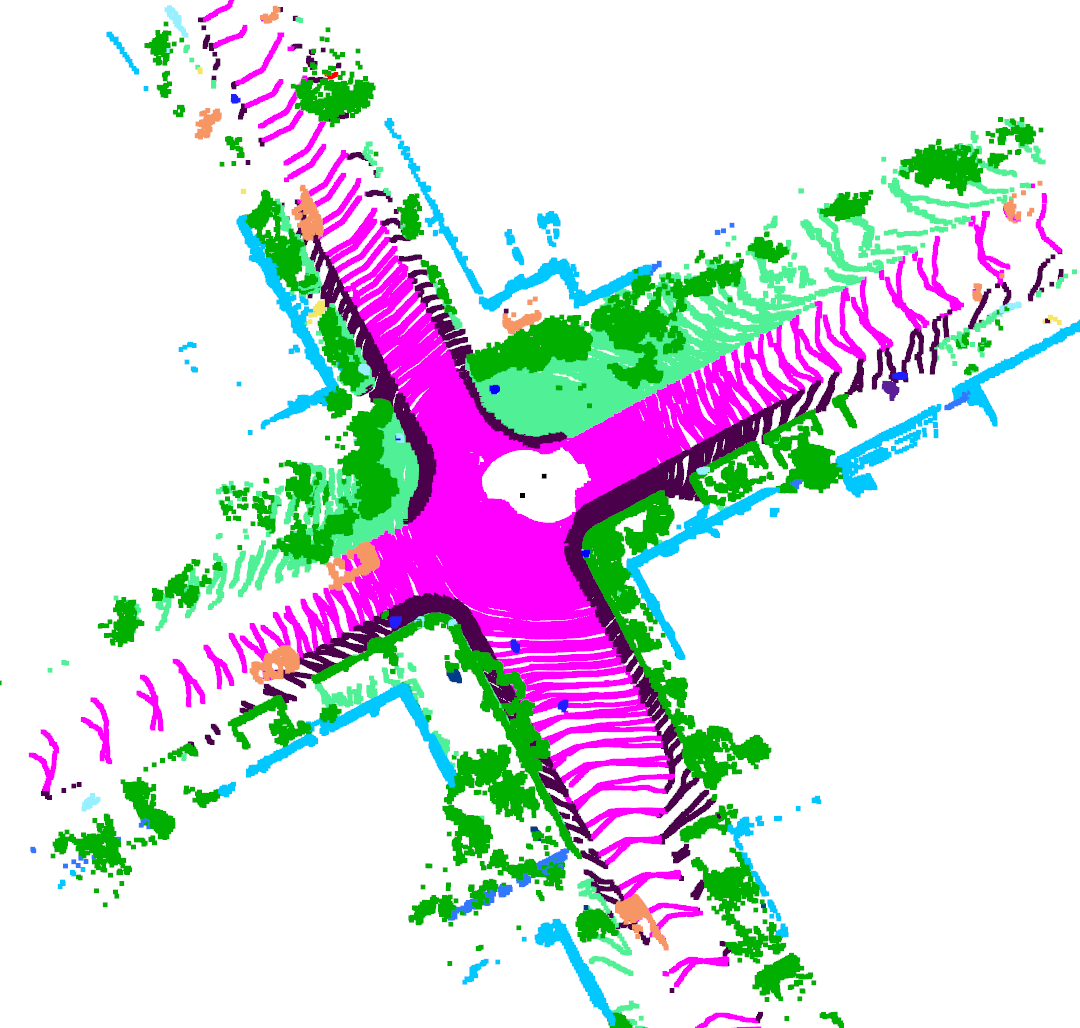}
        }
    \caption{An illustration of the two-step global semantic ICP.}
    \label{pic:yaw_icp}
 \end{figure}

\subsection{Global Semantic ICP}
It is known that the general ICP algorithm based on local iterative optimization is susceptible to local minimums~\cite{goicp}. For place recognition, we usually cannot get a valid initial value, which leads to the failure of the general ICP algorithm. To solve this, we propose the two-step global semantic ICP algorithm consisting of Fast Yaw Angle Calculate and Fast Semantic ICP. Benefited from the use of semantic information, our algorithm does not require any initial values to get satisfactory results.

\textbf{Fast Yaw Angle Calculate.}\label{ssc:yaw}
For scan context based methods, columns of their descriptor represent the yaw angle. The pure rotation of the LiDAR in the horizontal plane will cause the column shift of their descriptor. Scan context and Intensity Scan Context get the similarity score and the yaw angle at the same time. Specifically, they calculate similarity (or distance) with all possible column-shifted descriptors and find the maximum similarity (or minimum distance). However, there are two main disadvantages. Firstly, it's inefficient to compare the whole 2D descriptors by shifting. Secondly, they still try to get the maximum score for point clouds from different places (not loop closure). This obviously makes it more prone to false positives. To draw the above issues, we propose the semantic-based fast yaw angle calculate method.

Given a point cloud pair \( (P_1,P_2)\), we select representative objects such as buildings, tree trunks, and traffic signs based on semantic information. Then we convert the filtered clouds to polar coordinate in  the x-y plane:

\begin{equation}
    \begin{aligned}
    p_i&=[r_i,\varphi_i  ,x_i,y_i,\eta_i ] \\
    r_i&=\sqrt{x_i^2+y_i^2} \\
    \varphi_i &=arctan(\frac{y_i}{x_i} )
    \end{aligned}
\end{equation}
where \(p_i\) is the \(i^{th}\) point in each converted cloud, \(r_i\) and \(\varphi_i\) represent polar diameter and polar angle, respectively. Each converted cloud is then segmented to \( N_a\) sectors by yaw angle. We only keep the point with the smallest polar diameter in each sector. Finally, we get two clouds \(P_{I1}\) and \(P_{I2}\), with \(N_a\) elements. We sort the points in \(P_{I1}\) and \(P_{I2}\) according to the azimuth angle and save their corresponding polar diameters as vectors \(R_1\) and \(R_2\). Similar to the scan context, the shift of the column vector is related to the yaw angle:
 \begin{equation}\label{equ:shift}
    \begin{aligned}
     shift&=\mathop{argmin}\limits_{i, i\in [0,N_a]}\varPsi (R_1,R_2^i) \\
     \theta&=360-\frac{360\times shift}{N_a} 
    \end{aligned}
 \end{equation}
where \(R_2^i\) is \(R_2\) shifted by \(i^{th}\) element and \(\varPsi\) is defined as:
\begin{equation}
    \varPsi (R_1,R_2^i)=\left\lVert R_1-R_2^i\right\rVert_1 
\end{equation}

Compared with Scan Context and Intensity Scan Context, our method only needs to compare one-dimensional vectors; therefore, it is more efficient. Moreover, our method does not obtain the angle via maximizing the score, which is helpful to identify non-loop-closure point-cloud pairs. Fig.~\ref{pic:yaw_icp} shows the result of Fast Yaw Angle Calculate.

\begin{figure}[t]
    \centering
        \includegraphics[width=0.99\columnwidth]{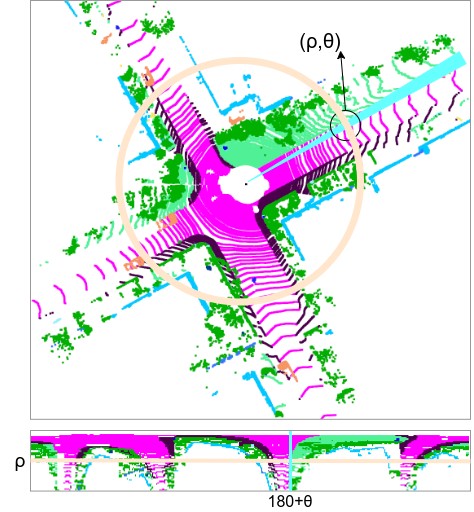}\vspace{-2mm}
    \caption{An example of generating SSC. \(\rho\) and \(\theta\) represent the polar diameter and polar angle, respectively. A sector corresponds to a descriptor column, while a ring corresponds to a row of the descriptor.}
    \label{pic:ssc}
 \end{figure}

\textbf{Fast Semantic ICP.}\label{ssc:icp}
Though most works ignore translation between point clouds, ignoring the translation causes considerable declines in our experiments. In fact, for methods based on scan context, translation will affect both the row and column of the descriptor. We can’t get the best result just by the column-shifted descriptor. Therefore, we propose a fast semantic ICP algorithm to correct the translation between point clouds.


To find the relative translation, we firstly rotate \(P_{I2}\) to the same direction as \(P_{I1}\), and the rotated point cloud is \(P_{Ia}\), which is defined as:
 \begin{equation}
     \begin{aligned}
         x_{ai}&=x_icos(\theta)-y_isin(\theta)\\
         y_{ai}&=x_isin(\theta)+y_icos(\theta)
     \end{aligned}
 \end{equation}
where \((x_i,y_i)\) and \((x_{ai},y_{ai})\) represent the \(i^{th}\) point in \(P_{I2}\) and \(P_{Ia}\) respectively. Our ICP problem can be defined as:
 \begin{equation}\label{eq:icp}
     \begin{aligned}
        (\Delta x,\Delta y)&=\mathop{argmin}\limits_{\Delta x,\Delta y}L=\mathop{argmin}\limits_{\Delta x,\Delta y}\sum_{i = 1}^{N_a}\Gamma(\eta_{ai},\eta_{ri})\\
        \times&\frac{(x_{ai}+\Delta x-x_{ri})^2+(y_{ai}+\Delta y-y_{ri})^2}{2}
     \end{aligned}
 \end{equation}
 where \((x_{ri},y_{ri})\) represents the corresponding point of \((x_{ai},y_{ai})\), which is the point closest to \((x_{ai},y_{ai})\) in \(P_{I1}\), $\eta_{ai}$ and $\eta_{ri}$ are semantic labels of the points. If $\eta_{ai}$ is equal to $\eta_{ri}$, then the output of $\Gamma(\eta_{ai},\eta_{ri})$ is $1$; otherwise, $0$. As our point clouds are ordered, we can search for the corresponding points near the position where the yaw angle is consistent with the target point. Specifically, our search interval for the \(i^{th}\) target point is:
 \begin{equation}
     [i+shift-\frac{N_l}{2},i+shift+\frac{N_l}{2}]
 \end{equation}
 where \(N_l\) is the length of search interval and \(shift\) is defined in Eq.~\ref{equ:shift}. After a certain number of iterations, we can get the relative translation between the input point clouds, shown in Fig.~\ref{pic:yaw_icp}.

\begin{figure*}[htb]
    \centering
        \subfigure[00]{
        \centering
        \includegraphics[width=0.64\columnwidth]{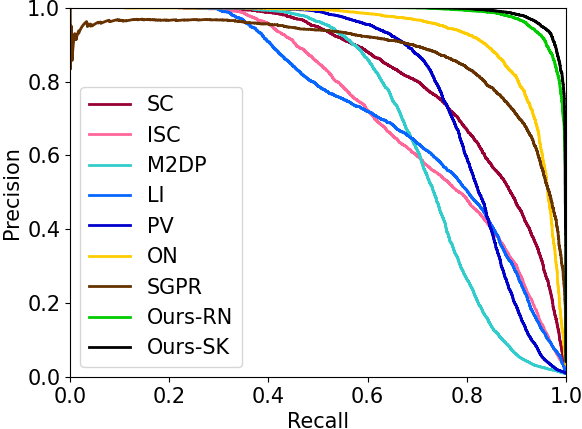}
        }
        \subfigure[02]{
        \centering
        \includegraphics[width=0.64\columnwidth]{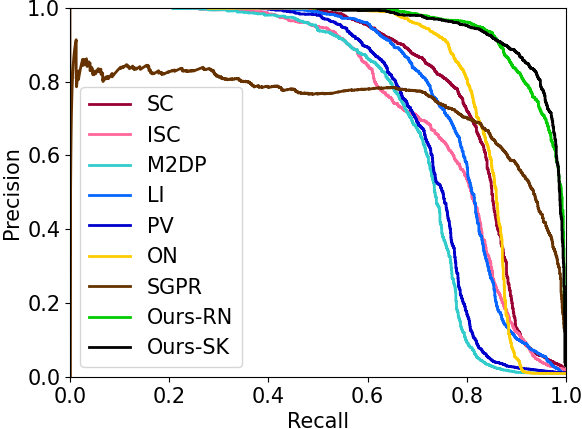}
        }
        \subfigure[05]{
        \centering
        \includegraphics[width=0.64\columnwidth]{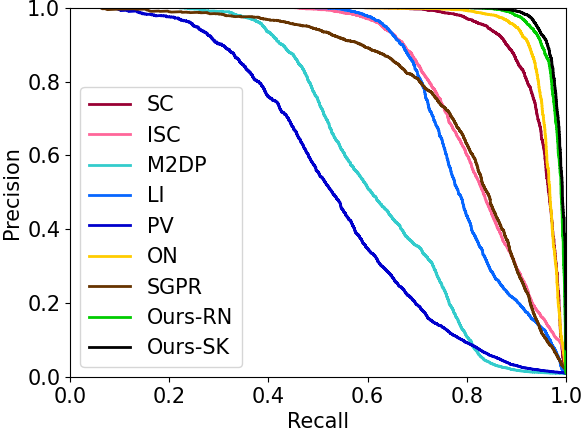}
        }\vspace{-3mm}
        \subfigure[06]{
        \centering
        \includegraphics[width=0.64\columnwidth]{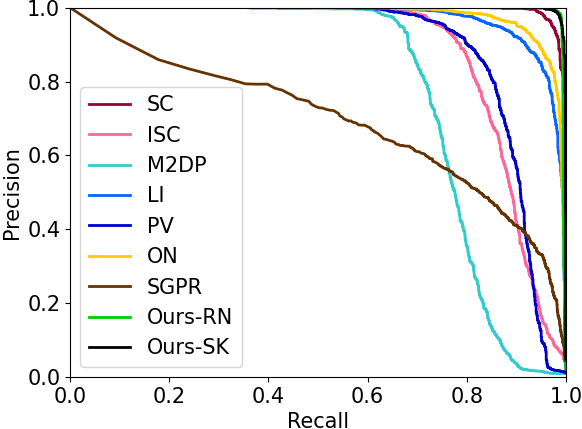}
        }
        \subfigure[07]{
        \centering
        \includegraphics[width=0.64\columnwidth]{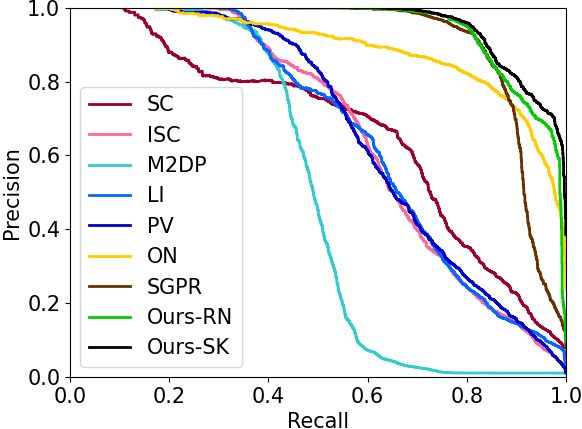}
        }
        \subfigure[08]{
        \centering
        \includegraphics[width=0.64\columnwidth]{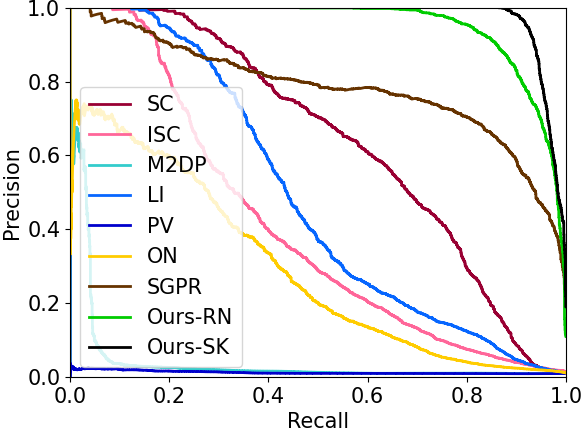}
        }\vspace{-3mm}
    \caption{Precision-Recall curves on KITTI dataset.}
    \label{pic:pr_curvs}
 \end{figure*}
 
  \begin{table*}[htb]\footnotesize
    \caption{\centering \(F_1\) max scores and Extended Precision on KITTI dataset}\vspace{-3mm}
    \label{table:F1}
    \begin{center}
    \begin{threeparttable}
        {
    \begin{tabular}{c c c c c c c c}
    \hline
    Methods & 00 & 02 & 05 & 06 & 07 &08&Mean\\ 
    \hline
    SC\cite{SC} & 0.750/0.609 & 0.782/0.632 &0.895/0.797&0.968/0.924&0.662/0.554&0.607/0.569&0.777/0.681\\
    ISC\cite{ISC}&0.657/0.627&0.705/0.613&0.771/0.727&0.842/0.816&0.636/0.638&0.408/0.543&0.670/0.661\\
    M2DP\cite{M2DP}&0.708/0.616&0.717/0.603&0.602/0.611&0.787/0.681&0.560/0.586&0.073/0.500&0.575/0.600\\
    LI\cite{LI}&0.668/0.626&0.762/0.666&0.768/0.747&0.913/0.791&0.629/0.651&0.478/0.562&0.703/0.674\\
    PV\cite{PV}&0.779/0.641&0.727/0.691&0.541/0.536&0.852/0.767&0.631/0.591&0.037/0.500&0.595/0.621\\
    ON\cite{ON}&0.869/0.555&0.827/0.639&0.924/0.796&0.930/0.744&0.818/0.586&0.374/0.500&0.790/0.637\\
    SGPR\cite{SGPR}&0.820/0.500&0.751/0.500&0.751/0.531&0.655/0.500&0.868/0.721&0.750/0.520&0.766/0.545\\
    Ours-RN&\underline{0.939}/\underline{0.826}&\underline{0.890}/\underline{0.745}&\underline{0.941}/\underline{0.900}&\textbf{0.986}/\textbf{0.973}&\underline{0.870}/\underline{0.773}&\underline{0.881}/\underline{0.732}&\underline{0.918}/\underline{0.825}\\
    Ours-SK&\textbf{0.951}/\textbf{0.849}&\textbf{0.891}/\textbf{0.748}&\textbf{0.951}/\textbf{0.903}&\underline{0.985}/\underline{0.969}&\textbf{0.875}/\textbf{0.805}&\textbf{0.940}/\textbf{0.932}&\textbf{0.932}/\textbf{0.868}\\
    \hline
    \end{tabular}
    }
    \begin{tablenotes} 
        \footnotesize
        \item \(F_1\) max scores and Extended Precision: \(F_1\) max scores / Extended Precision. The best scores are marked in bold and the second best scores are underlined.
     \end{tablenotes}
    \end{threeparttable}
    \end{center}
    \end{table*}

\subsection{Semantic Scan Context}
Scan Context and Intensity Scan Context uses the points' height and reflection intensity as features, respectively. Their methods essentially take advantage of the different characteristics of different objects in the scene. However, height and reflection intensity is only low-level features of the object which are not representative enough. We explore to use the high-level semantic features to represent scenes and thus propose the Semantic Scan Context descriptor.

\smallskip\noindent\textbf{Descriptor definition.} Given a point cloud \(P\), we first convert it to the polar coordinate system as we did in Section~\ref{ssc:yaw}. Then, like scan context, we divide the point cloud into \(N_s\times N_r\) blocks along the azimuthal and radial directions. Each block is represented by:
\begin{equation}
    \begin{aligned}
        B_{ij}=\{\eta_k|\frac{(i-1)\cdot R_{max}}{N_r}\leq r_k< \frac{i\cdot R_{max}}{N_r},\\
        \frac{(j-1)\cdot 2\pi }{N_s}-\pi \leq \varphi_k<\frac{j\cdot 2\pi }{N_s}-\pi\}
    \end{aligned}
\end{equation}
where \(R_{max}\) is the the maximum effective measurement distance of LiDAR, \(i\in [1,N_r]\) and \(j\in [1,N_s]\). Our descriptor can be defined by:
\begin{equation}
    S(i,j)=f(B_{ij})=\mathop{argmax}\limits_{\eta \in B_{ij}} E(\eta )\label{eq:descriptor}
\end{equation}
\(f\) is an encoding function to encode features of \(B_{ij}\). Note that if \(B_{ij}=\varnothing ,~f(B_{ij})=0\). We manually set the priority of different semantics in function \(E \) to show their representativeness. We believe objects that appear less frequently in the scene are more representative (e.g., traffic signs are more representative than roads). 

\smallskip\noindent\textbf{Similarity Scoring.} Given aligned clouds \(P_1\) and \(P_{a}\), we can get their descriptors \(S_1\) and \(S_2\) by Eq.~\ref{eq:descriptor}. Then the similarity score between them can be calculated by:
\begin{small}
\begin{equation}
    \begin{aligned}
        score=\frac{\sum\limits_{ 1\leq i\leq N_r}\sum\limits_{1\leq j\leq N_s} I(S_1(i,j)=S_2(i,j))}{\sum\limits_{1\leq i\leq N_r}\sum\limits_{1\leq j\leq N_s} I(S_1(i,j)\neq 0~or~S_2(i,j)\neq 0)}
    \end{aligned}
\end{equation}
\end{small}
where \(I\) is the indicator function, defined by:
\begin{equation}
    I(x)=\left\{
    \begin{aligned}
        &1& & x~is~true\\
        &0& & x~is~false 
    \end{aligned}
    \right.
\end{equation}
Fig.~\ref{pic:ssc} shows Semantic Scan Context creation.

    \begin{figure*}[htb]
        \centering
            \subfigure[Average \(F_1\) max scores]{
            \centering
            \includegraphics[width=0.95\columnwidth]{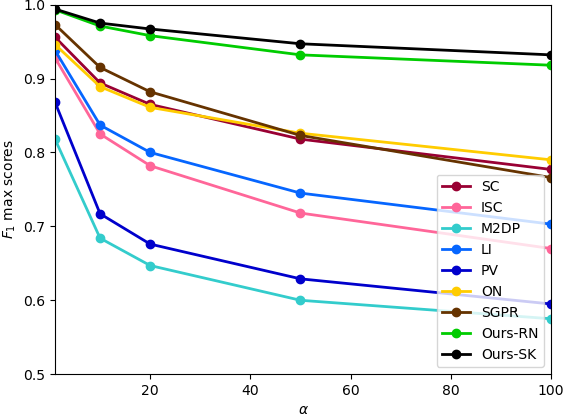}
            }
            \subfigure[Average \(EP\)]{
            \centering
            \includegraphics[width=0.95\columnwidth]{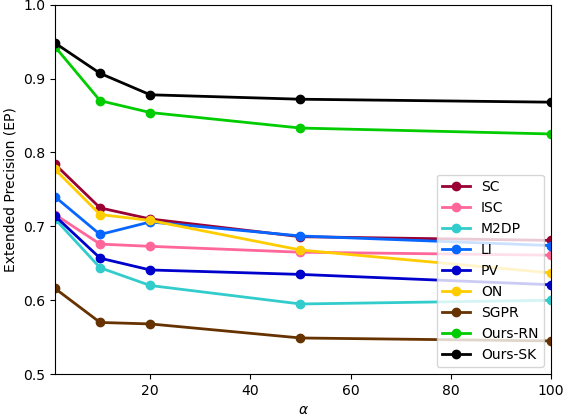}
            }\vspace{-3mm}
        \caption{Average \(F_1\) max score and Average Extended Precision corresponding to different \(\alpha\).}
        \label{pic:F1}
     \end{figure*}

\section{EXPERIMENTS}

\subsection{Experiment Setup}\label{exp:setup}
We conduct experiments on the KITTI odometry dataset\cite{kitti} collected by a 64-ring LiDAR, which contains 11 training sequences (00-10) with ground truth poses. We choose sequences with loop-closure (00,02,05,06,07,08) for evaluation and note that sequence 08 has reverse loops while others are in the same direction. Similar to SGPR\cite{SGPR}, we regard the point cloud pair with a relative distance less (greater) than 3m (20m) as a positive (negative) sample. Since there are too many negative samples, we only select a part of the negative samples for evaluation. Specifically, if there are \(N_{p}\) positive samples in a sequence, we will randomly select \(\alpha \cdot N_{p}\) negative samples. We can adjust the proportion of negative samples by changing the coefficient \(\alpha\).

The ground-truth semantic labels are from the SemanticKITTI dataset\cite{semkitti}. We also test our method with the semantic segmentation algorithm (RangeNet++~\cite{rangenet++})  to prove that our method can be applied to noisy predictions in real situations. In our experiments, we set \(N_a=360,~N_l=20,~N_s=360,~N_r=50\). All experiments are done on the same system with an Intel i7-9750H @3.00GHz CPU with 16 GB RAM.

\subsection{Place Recognition Performance}\label{seu:pr}
As mentioned in Section~\ref{exp:setup}, we use both ground-truth semantic labels (Ours-SK) and predicted semantic labels (Ours-RN) for testing. We compare our approach with the state-of-the-art methods, including Scan Context\cite{SC} (SC), Intensity Scan Context\cite{ISC} (ISC), M2DP\cite{M2DP}, LiDAR Iris\cite{LI} (LI), PointNetVLAD\cite{PV} (PV), OverlapNet\cite{ON} (ON), and SGPR\cite{SGPR}. For SGPR, we use their pre-trained models trained with the 1-fold strategy. As we cannot reproduce the results of OverlapNet, we use the pre-trained model provided by the author. The model is trained on sequences 03-10, so sequences 05, 06, 07, 08 are included in the training set.

\textbf{Fixed \(\alpha\)}. In this experiment, we set \(\alpha\) to 100, which means the number of negative samples is \(100N_{p}\). Fig.~\ref{pic:pr_curvs} shows the precision-recall curve of each method. Additionally, we also use the maximum \(F_1\) score and Extended Precision\cite{EP} (EP) shown in Tab.~\ref{table:F1} to analyze the performance. The \(F_1\) score is defined as:
\begin{equation}
    F_1=2\times \frac{P\times R}{P+R}
\end{equation}
where \(P\) and \(R\) represent the Precision and Recall, respectively; \(F_1\) is the harmonic mean of \(P\) and \(R\). It treats \(P\) and \(R\) as equally important and measures the overall performance of classification. The Extended Precision is defined as:
\begin{equation}
    EP=\frac{1}{2}(P_{R0}+R_{P100})
\end{equation}
where \(P_{R0}\) is the precision at minimum recall, and \(R_{P100}\) is the max recall at \(100\%\) precision. \(EP\) is specifically designed metrics for place recognition algorithms. 

As shown in Fig.~\ref{pic:pr_curvs} and Tab.~\ref{table:F1}, Ours-SK surpasses other methods in all indicators of all sequences with a large margin. Especially in sequence 08, which has only reverse loops, the performance of other methods drops significantly while our method still performs well. This indicates that our method is robust to view angle changes. OverlapNet performs well on most sequences except 08. We guess this is because it uses the normal of the point cloud, which will change as the point cloud rotates. Therefore, this method cannot robustly handle reverse loops. SGPR works well on indicator the \(F_1\) max score but poorly on the Extended Precision. We find that it gives some negative samples a huge score, which causes the recall to be almost zero when the accuracy reaches \(100\%\). The result of Ours-RN is slightly worse than Ours-SK as expected. As the difference is not obvious, it means that our approach can adapt to semantic segmentation algorithms for actual systems.

\textbf{Change \(\alpha\)}. In this experiment, we change the value of \(\alpha\) to analyze the influence of the number of negative samples on those algorithms. Fig.~\ref{pic:F1} shows the Average \(F_1\) max score and Average Extended Precision corresponding to different \(\alpha\). It clearly shows that our method performs better than others no matter how much \(\alpha\) is taken. As \(\alpha\) increases, the performance of all methods gradually decreases, but our method is less affected, showing that our method can effectively identify negative samples. For place recognition, negative samples are generally far more than positive samples, which is one key reason why our method leads in metrics far ahead. Moreover, identifying negative samples is significant as false positives will bring fatal crashes to the SLAM system.

    \begin{table}[t]\footnotesize
        \caption{\centering Yaw error on KITTI dataset}\vspace{-3mm}
        \label{table:yaw}
        \begin{center}
        \begin{threeparttable}
            {
        \begin{tabular}{c c c c c c c c}
        \hline
        sequences & SC (deg) & ISC (deg) & ON (deg) & Ours-SK (deg) \\ 
        \hline
        00 & 11.526&\textbf{0.829}&2.595&0.891\\
        02 &11.301&1.343&4.911&\textbf{1.142}\\
        05 &18.394&0.904&3.329&\textbf{0.653}\\
        06 &4.074&\textbf{0.534}&1.124&0.759\\
        07 &21.862&0.684&2.233&\textbf{0.512}\\
        08 &49.170&3.856&68.622&\textbf{1.878}\\
        Average &19.388&1.358&13.802&\textbf{0.973}\\
        \hline
        \end{tabular}
        }
        \end{threeparttable}
        \end{center}
        \end{table}
     
\subsection{Pose Accuracy}
As described in Section~\ref{ssc:icp}, our approach can estimate the 3D relative pose \( (\Delta x, \Delta y, \theta)\), while most other methods cannot estimate pose or can only estimate 1D pose (yaw). We compare our method with Scan Context, Intensity Scan Context, and Overlap. The ground-truth pose is calculated by:
\begin{equation}
    \begin{aligned}
    T&=T_1^{-1}T_2\\
    (\Delta x,\Delta y,\theta)&=(T(1,3),T(2,3),arctan(\frac{T(2,1)}{T(1,1)}))
    \end{aligned}
\end{equation}
where \(T_1\in SE(3)\) and \(T_2\in SE(3)\) represent the pose of \(P^1\) and \(P^2\), respectively. Since the pitch and roll angles are hardly changed in autonomous vehicles, we ignore them.

Tab.~\ref{table:yaw} shows the relative yaw error on the KITTI dataset. We can see that our method outperforms other methods in terms of the average relative yaw error. Especially in the challenging sequence 08, affected by the reverse loop, most methods perform poorly, while our method can still accurately estimate the yaw angle. This again shows that our method can handle the reverse loop well. As mentioned in Section~\ref{seu:pr}, OverlapNet performs poorly due to its inability to handle reverse loops.

\begin{figure}[t]
        \centering
            \includegraphics[width=0.95\columnwidth]{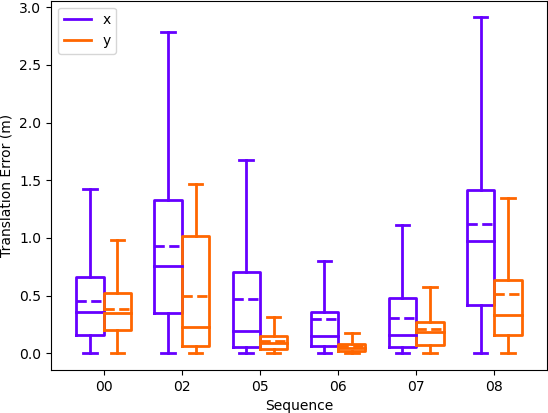}
        \caption{Translation error.}
        \label{pic:trans}
     \end{figure}

Fig.~\ref{pic:trans} shows the relative translation error of our approach on the KITTI dataset. As shown, our method can estimate accurate relative translation, which is currently not possible with other methods to our knowledge. Thus, our Fast Yaw Angle Calculate and Fast Semantic ICP approaches can give accurate 3D pose estimation. This can provide a good initial value for the ICP algorithm to obtain a 6D pose or directly serve as a global constraint in the SLAM system.



\begin{table}[t]\footnotesize
    \caption{\centering Contribution of individual components}\vspace{-3mm}
    \label{table:Ablation}
    \begin{center}
    \begin{tabular}{c c c c c c c c}
    \hline
    Yaw & ICP & Semantic & \(F_1/EP\)&Decrease\\ 
    \hline
    ~&$\surd$&$\surd$&0.896/0.820&3.6\%/4.8\%\\
    $\surd$&~&$\surd$&0.757/0.685&17.5\%/18.3\%\\
    $\surd$&$\surd$&~&0.775/0.762&15.7\%/10.6\%\\
    $\surd$&$\surd$&$\surd$&0.932/0.868&0.0\%/0.0\%\\
    \hline
    \end{tabular}
    \end{center}
    \end{table}

\subsection{Ablation Study}
We design an ablation study to investigate the contribution of each component. Specifically, we remove or replace a module at a time and then calculate the \(F_1\) max scores and Extended Precision. To show the contribution of our Fast Yaw Angle Calculate method, we replace this module with the method used in scan context -- shift the column of descriptors and calculate the maximum similarity score while obtaining the yaw angle. Similarly, we replace the semantic label in the descriptor by maximum \(z\) to see semantic contribution. To evaluate the contribution of our Fast Semantic ICP approach, we directly set \(\Delta x\) and \(\Delta y\) to 0. As shown in Tab.~\ref{table:Ablation}, after removing Yaw, ICP, and Semantic, the average \(F_1\) max score decrease by \(3.6\%\), \(17.5\%\), \(15.7\%\), and the average Extended Precision decrease by \(4.8\%\), \(18.3\%\), \(10.6\%\). Therefore, the following conclusions can be drawn:
\begin{itemize}
     \item Compared with other methods, our approach can get a more accurate yaw angle and translation.
     \item As we emphasized, the small translation has a significant impact on scan context-based methods. Simply ignoring the translation will greatly weaken the performance.
     \item High-level features, like semantics, can bring considerable improvements in the scene description.
\end{itemize}

\begin{table}[t]\footnotesize
    \caption{\centering Average time cost on KITTI 08}\vspace{-3mm}
    \label{table:time}
    \begin{center}
    \begin{threeparttable}
    {
    \begin{tabular}{c c c c c c c}
    \hline
    Methods &Size& Description & Retrieval & ICP & Total\\ 
    \hline
    SC&$20\times 60$&4.825&0.158&-&4.983\\
    ISC&$20\times 90$&3.094&0.800&-&\textbf{3.894}\\
    Ours&$50\times 360$&\textbf{2.563}&\textbf{0.066}&2.126&4.755\\
    \hline
    \end{tabular}
    }
    \begin{tablenotes} 
            \footnotesize
            \item The unit of time in the table is milliseconds.
         \end{tablenotes}
    \end{threeparttable}
    \end{center}
    \end{table}

\subsection{Efficiency}
To evaluate the efficiency, we set $\alpha$ to $1$ and compare the average time cost of our method with Scan Context and Intensity Scan Context on sequence 08. As shown in Tab.~\ref{table:time}, the total time cost of our approach is acceptable. As we use the obtained 3D pose to align the point clouds in advance, we don't need to shift the column of descriptors during the matching stage, so our retrieval speed is extremely fast. Our two-step global semantic ICP only takes 2.126 milliseconds on average. This algorithm is fast due to the following reasons. Firstly, since we only keep \(N_a\) (360 taken in our experiments) points, the computational cost is greatly reduced compared to the original point cloud (about 120,000 points). Secondly, We divide the algorithm into two steps, first calculate the yaw angle, and then iteratively calculate $\Delta x$ and $\Delta y$, which simplifies the algorithm and speeds up the calculation. Thirdly, when calculating $\Delta x$ and $\Delta y$, we use the yaw angle to align the input clouds in advance. Therefore we don’t need to traverse the entire point cloud when looking for the corresponding points. Instead, we can find them near the corresponding positions, which greatly reduces the number of searches.

\section{CONCLUSION}
In this paper, we propose a novel semantic-based global descriptor for place recognition. We propose a two-step global semantic ICP to obtain the 3D pose \((x,y,yaw)\) of the point cloud pair, aligning the point clouds to improve the descriptor matching accuracy. In addition, it can provide good initial values for point cloud registration. We achieve leading performance on the KITTI odometry dataset compared to the state-of-the-art methods. 

Our method also has some limitations. Like most place recognition methods, our method does not consider pitch angle and roll angle. Therefore, our method may fail in some extreme scenarios.

In the future work, we will try to solve the above problems and further explore the application of semantic information in LiDAR-based SLAM systems.


\bibliographystyle{ieeetr}
\bibliography{main}

\end{document}